\documentclass[10pt,twocolumn,letterpaper]{article}

\usepackage{cvpr}
\usepackage{times}
\usepackage{epsfig}
\usepackage{graphicx}
\usepackage{amsmath}
\usepackage{amssymb}
\usepackage{xcolor}

\usepackage[breaklinks=true,bookmarks=false]{hyperref}

\cvprfinalcopy %

\newcommand{\parag}[1]{\vspace{-3mm}\paragraph{#1}}

\definecolor{orange}{rgb}{1,0.5,0}
\definecolor{deeppink}{RGB}{255,20,147}

\newif\ifdraft
\draftfalse

\ifdraft

\else

\fi

\newcommand{\dronepose}{\mathbf{D}}
\newcommand{\openpose}{\mathbf{M}}
\newcommand{\liftpose}{\mathbf{L}}
\newcommand{\pose}{\mathbf{\Theta}}
\newcommand{\posevec}{\mathbf{\Theta}^{\text{c}}} %
\newcommand{\bone}{\mathbf{b}}
\newcommand{\bonecalib}{\mathbf{b_\text{calib}}}

\newcommand{\para}[1]{\textbf{#1}} %

\newcommand{\R}{\mathbb{R}} %

\def\cvprPaperID{9312} %

\ifcvprfinal\fi

\begin{document}
\renewcommand\footnotemark{}
\renewcommand\footnoterule{}

\title{ActiveMoCap: Optimized Viewpoint Selection \\for Active Human Motion Capture}

\author{Sena Kiciroglu$^{1}$ \,\, Helge Rhodin$^{1,2}$ \,\, Sudipta N. Sinha$^{3}$ \,\, Mathieu Salzmann$^{1}$ \,\, Pascal Fua$^{1}$\\\vspace{-2pt}
	$^1$ CVLAB, EPFL \qquad $^2$ Imager Lab, UBC \qquad $^3$ Microsoft  \vspace*{1.5ex}
}

\maketitle

\begin{abstract}
The accuracy of monocular 3D human pose estimation depends on the viewpoint from which the image is captured. While freely moving cameras, such as on drones, provide control over this viewpoint, automatically positioning them at the location which will yield the highest accuracy remains an open problem. This is the problem that we address in this paper. Specifically, given a short video sequence, we introduce an algorithm that predicts which viewpoints should be chosen to capture future frames so as to maximize 3D human pose estimation accuracy. The key idea underlying our approach is a method to estimate the uncertainty of the 3D body pose estimates. We integrate several sources of
uncertainty, originating from deep learning based regressors and temporal smoothness. Our motion planner yields improved 3D body pose estimates and outperforms or matches existing ones that are based on person following and orbiting.

\end{abstract}

\section{Introduction}

\begin{figure}[t!]
    \centering
    \includegraphics[width=0.45\textwidth]{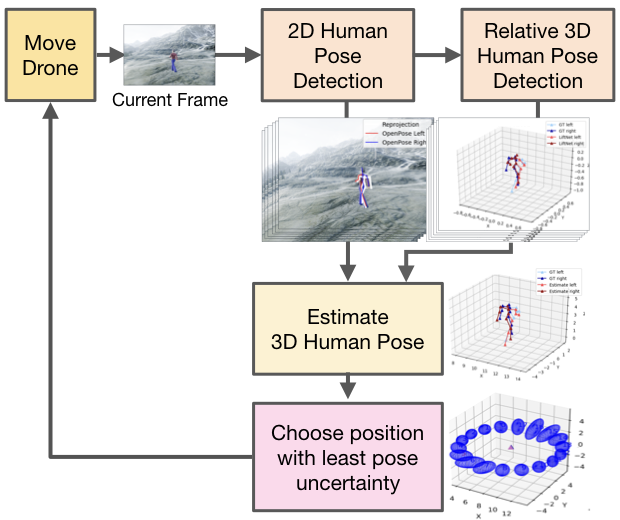}
    \caption{\textbf{Method overview.} The 2D and 3D human pose is inferred from the current frame of the drone footage, using off the shelf CNNs. The 2D pose and relative 3D pose of the last $k$ frames is then used to optimize for the global 3D human motion. The next view of the drone is chosen so that the uncertainty of the human pose estimation from that view is minimized, which improves reconstruction accuracy.}
    \label{fig:drone}
\end{figure}

Monocular approaches for 3D human pose estimation have improved significantly in recent years, but their accuracy remains relatively low. In this paper, we explore the use of a moving camera whose motion we can control to resolve ambiguities inherent to monocular 3D reconstruction and to increase pose estimation accuracy. This is known as {\it active vision} and has received surprisingly little attention in the context of using modern approaches to body pose estimation. An active motion capture system, such as one based on a personal drone, would allow one to film themselves performing a physical activity and analyze their motion, for example to get feedback on their performance. When using only one camera, the quality of such feedback will strongly depend on selecting the most beneficial viewpoints for pose estimation. Fig.~\ref{fig:drone} depicts an overview of our approach based on a drone-based monocular camera.

In this paper, we introduce an algorithm designed to continuously position a moving camera at optimal viewpoints to maximize the 3D pose estimation accuracy for a freely moving subject. We achieve this by moving the camera in 6D pose space to viewpoints that maximize a utility function designed to predict reconstruction accuracy. However, the utility function cannot be defined in terms of reconstruction accuracy because doing so would require knowing the true person and camera position, leading to a chicken and egg problem. Instead we use prediction uncertainty as a surrogate for accuracy. This is a common strategy used in robot navigation systems for unknown scenes where the robot explores areas that are most incomplete in its internal map representation~\cite{Palazzolo17}. However, in our situation, estimating uncertainty is much more difficult since multiple sources of uncertainty need to be considered. These include uncertainties about what the subject will do next, the reliability of the pose estimation algorithm, and the accuracy of distance estimation along the camera's line of sight. 

Our key contribution is therefore a formal model that provides an estimate of the \emph{posterior variance} and probabilistically fuses these sources of uncertainty with appropriate prior distributions. This has enabled us to develop an active motion capture technique that takes raw video footage as input from a moving aerial camera and continuously computes future target viewpoints for positioning the camera, in a way that is optimized for human motion capture. We demonstrate our algorithm in two different scenarios and compare it against standard heuristics, such as constantly rotating around the subject and maintaining a constant angle with respect to the subject. We find that when allowed to choose the next viewpoint without physical constraints, our algorithm outperforms the baselines consistently. For simulated drone flight, our results are on par with constant rotation, which we conclude is the best trajectory to choose in the case of no obstacles blocking the circular flight path. Our code is available at \url{https://github.com/senakicir/ActiveMoCap}

\section{Related work}
\label{sec:related}

Most recent approaches to markerless motion capture rely on deep networks that regress 3D pose from monocular images ~\cite{Martinez17,Mehta17a,Pavlakos16,Tome17,Popa17,Rogez17,Pavlakos17,Zhou17d,Tekin17a,Sun17,Zanfir18a,Xiang19,Kanazawa19}. While a few of these methods improve robustness by enforcing temporal consistency~\cite{Pavllo19}, none considers the effect that actively controlling the camera may have on accuracy. The methods most closely related to ours are therefore those that optimize camera placement in multi-camera setups and those that guide robots in a previously-unknown environment. 

\parag{Optimal Camera Placement for Motion Capture.}

Optimal camera placement is a well-studied problem in the context of static multi-view setups. Existing solutions rely on maximizing image resolution while minimizing self-occlusion of body parts~\cite{Chen00b,Aissaoui18} or target point occlusion and triangulation errors~\cite{Rahimian16}. However, these methods operate offline and on pre-recorded exemplar motions. This makes them unsuitable for motion capture using a single moving camera that films {\it a priori} unknown motions in a much larger scene where estimation noise can be high.

In \cite{Pirinen19} multiple cameras poses are optimized for triangulation of joints in a dome environment using a self-supervised reinforcement learning approach. In our case, we consider the monocular problem. Our method is not learning based, we try to obtain the next best view from the loss function itself.

\parag{View Planning for Static and People Reconstruction.}

There has been much robotics work on active reconstruction and view planning. This usually involves moving so as to maximize information gain while minimizing motion cost,  for example by a discretizing space into a volumetric grid and counting previously unseen voxels~\cite{Isler16, Daudelin17} or by accumulating estimation uncertainty~\cite{Palazzolo17}. When a coarse scene model is available, an optimal trajectory can be found using offline optimization~\cite{Roberts17,Hepp18}.  This has also been done to achieve desired aesthetic properties in cinematography~\cite{Gebhardt18}. Another approach is to use reinforcement learning to define policies~\cite{Choudhury17} or to learn a metric~\cite{Hepp18b} for later online path planning. These methods deal with rigid unchanging scenes, except the one in~\cite{Cheng18} that performs volumetric scanning of people during information gain maximization. However, this approach can only deal with very slowly moving people who stay where they are.  

\parag{Human Motion Capture on Drones.}

Drones can be viewed as flying cameras and are therefore natural targets for our approach. One problem, however, is that the drone must keep the person in its field of view. To achieve this, the algorithm of~\cite{Zhou18} uses 2D human pose estimation in a monocular video and non-rigid structure from motion to reconstruct the articulated 3D pose of a subject, while that of~\cite{Naegeli17} reacts online to the subject's motion to keep them in view and to optimize for screen-space framing objectives. AirCap~\cite{Nitin19} calculates trajectories of multiple drones that aim to keep the person in view while simultaneously performing object avoidance. This was extended in~\cite{Tallamraju19} so as to optimize multiple MAV trajectories by minimizing the uncertainty of the global human position.
In~\cite{Naegeli18}, this was integrated into an autonomous system that actively directs a swarm of drones and simultaneously reconstructs 3D human and drone poses from onboard cameras. This strategy implements a pre-defined policy to stay at constant distance to the subject and uses pre-defined view angles ($90^\circ$ between two drones) to maximize triangulation accuracy. This enables mobile large-scale motion capture, but relies on markers for accurate 2D pose estimation. In~\cite{Xu16b}, three drones are used for markerless motion capture, using an RGBD video input for tracking the subject.

In short, existing methods either optimize for drone placement but for mostly rigid scenes, or estimate 3D human pose but without optimizing the camera placement. \cite{Pirinen19} performs optimal camera placement for multiple cameras. Here, we propose an approach that aims to find the best next drone location for monocular view so as to maximize 3D human pose estimation accuracy.

\section{Active Human Motion Capture}

Our goal is to continuously position the camera in 6D pose space so that the acquired by the camera can be used to achieve the best overall human pose estimation accuracy. What makes this problem challenging is that, when we decide where to send the camera, we do not yet know where the subject will be and in what position exactly. We therefore have to guess. To this end, we propose the following three-step approach depicted by Fig.~\ref{fig:drone}:
\vspace{-0.3em}
\begin{enumerate}
 \itemsep-0.4em

 \item Estimate the 3D pose up to the current time instant. 
 
 \item Predict the person's future location and 3D pose at the time the camera acquires the next image, including an uncertainty estimate. 
 
 \item Selectthe optimal camera pose based on the uncertainty estimate and move the camera to that viewpoint.\end{enumerate}
\vspace{-0.3em}
We will consider two ways the camera can move. In the first case, the camera can teleport from one location to the next without restriction, allowing us to explore the theoretical limits of our approach. Such a teleportation mode can be simulated using a multi-camera setup, enabling us to evaluate our model on both simulated data and real image datasets acquired from multiple viewpoints. In the second, more realistic scenario, the camera is carried by a simulated drone, and we must take into account physical limits about the motion it can undertake. %
\subsection{3D Pose Estimation}\label{sec:pose_est_sec}

The 3D pose estimation step takes as input the video feed from the on-board camera over the past $N$ frames and outputs for each frame, $t \in (1, \dots, N)$, the 3D human pose, represented as 15 3D points $\pose^t \in \R^{15\times3}$, and the drone pose, as 3D position and rotation angles $\dronepose^t \in\R^{2\times3}$.
 Our focus is on estimating the 3D human pose using the real-time method proposed by \cite{Cao17}, which detects the 2D locations of the human's major joints in the image plane, $\openpose^t \in \R^{15\times2}$, and the subsequent use of \cite{Tekin17a}, which lifts these 2D predictions to 3D pose, $\liftpose^t \in \R^{15\times3}$. However, these per-frame estimates are error prone and relative to the camera.
To remedy this, we fuse 2D and 3D predictions with temporal smoothness and bone-length constraints in a space-time optimization. This exploits the fact that the drone is constantly moving so as to disambiguate the individual estimates. The bone lengths, $\bonecalib$, of the subject's skeleton are computed during an apriori calibration stage, where the subject has to stand still for $20$ seconds. This is performed only once for each subject. Formally, we optimize for the global 3D human pose by minimizing an objective function $E_\text{pose}$, which we detail below. 

\subsubsection{Formulation}
Our primary goal is to improve the global 3D human pose estimation of a subject changing position and pose. We optimize the time-varying pose trajectories across the last $k$ frames. Let $t$ be the last observed frame. We capture the trajectory of poses $\pose^{t-k}$ to $\pose^{t}$ in the pose matrix $\pose$.
We then write an energy function
\vspace{-.1cm}
\begin{align}
E_\text{pose}%(\pose, \openpose, \dronepose, \liftpose, \bone) 
&= E_\text{proj}(\pose, \openpose, \dronepose) +  E_\text{lift}(\pose, \liftpose) \nonumber\\
&+E_\text{smooth}(\pose) 
+E_\text{bone}(\pose, \bone) \;.
\end{align}
The individual terms are defined as follows.
The lift term, $E_\text{lift}$, leverages the 3D pose estimates, $\liftpose$, from LiftNet~\cite{Tekin17a}. Because these are relative to the hip and without absolute scale, we subtract the hip position from our absolute 3D pose, $\pose^t$, and apply a scale factor $m$ to $\liftpose$ to match the bone lengths $\bonecalib$ in the least-square sense. We write 
\vspace{-.1cm}
\begin{align}
E_\text{lift}(\pose, \liftpose) = \omega_l \sum_{i=t-k}^{t} \lVert m\cdot \liftpose^i - (\pose^i- \pose^i_\text{hip joint}) \rVert_2^2 \;,
%-\liftpose^i_\text{hip joint}
\end{align}
with $\omega_l$ its relative weight. 

The projection term measures the difference between the detected 2D joint locations and the projection of the estimated 3D pose in the least-square sense. We write it as
\vspace{-0.3cm}
\begin{align}
E_\text{proj}(\pose, \openpose, \dronepose) = \omega_p \sum_{i=t-k}^{t} \lVert \openpose^i- \Pi(\pose^i, \dronepose^i, \mathbf{K})\rVert_2^2 \;,
\end{align}

where $\Pi$ is the perspective projection function, $\mathbf{K}$ is the matrix of camera intrinsic parameters, and $\omega_p$ is a weight that controls the influence of this term.

The smoothness term exploits that we are using a continuous video feed and that the motion is smooth by penalizing velocity computed by finite differences as

\vspace{-0.5em}
\begin{align}
E_\text{smooth}(\pose) = \omega_s\sum_{i=t-k+1}^{t} \lVert( \pose^{i+1}-\pose^{i})\rVert_2^2  \;.
\end{align}
with $\omega_s$ as its weight.

To further constrain the solution space, we use our knowledge of the bone lengths $\bonecalib$ found during calibration and penalize deviations in length. The length of each bone $b$ in the set of all bones $b_\text{all}$ is found as $\bone^t_b = \lVert(\pose_{b_1}-\pose_{b_2})\rVert_2$ for frame $t$. The bone length term is then defined as
\vspace{-0.5em}
\begin{align}
E_\text{bone}(\pose) = \omega_b\sum_{i=t-k}^{t} \sum_{b \in b_\text{all}} d(\bone^i_b, \bonecalib_{,b}) \;,
\end{align}
with $\omega_b$ as its weight.

The complete energy $E_\text{pose}$ is minimized by gradient descent at the beginning of each control cycle, to get a pose estimate for control. The resulting pose estimate $\hat{\pose}$ is the maximum a posteriori estimate in a probabilistic view. 

\subsubsection{Calibration Mode}
Calibration mode only has to be run once for each subject to find the bone lengths, $\bonecalib$. In this mode, the subject is assumed to be stationary. The situation is equivalent to having the scene observed from multiple stationary cameras, such as in \cite{Rhodin16a}. We find the single static pose $\posevec$ that minimizes
\begin{equation}
E_\text{calib} = E_\text{proj}(\posevec, \openpose, \dronepose) + E_\text{symmetry}(\posevec).
\end{equation}
In this objective, the projection term, $E_\text{proj}$, is akin to the one in our main formulation but acts on all calibration frames. 
It can be written as
\begin{equation}
E_\text{proj}(\posevec, \openpose, \dronepose) = \omega_\text{p} \sum_{i=0}^{t}\lVert \openpose^i - \Pi(\posevec, \dronepose^i, \mathbf{K} )\rVert_2^2 \;,
\end{equation}
with $\omega_\text{p}$ controlling its influence.
The symmetry term, $E_\text{symmetry}$, ensures that the left and right limbs of the estimated skeleton have the same lengths by penalizing the squared difference of their lengths.

\subsection{Next Best View Selection}

\begin{figure}
	\centering
	\includegraphics[width=1\linewidth]{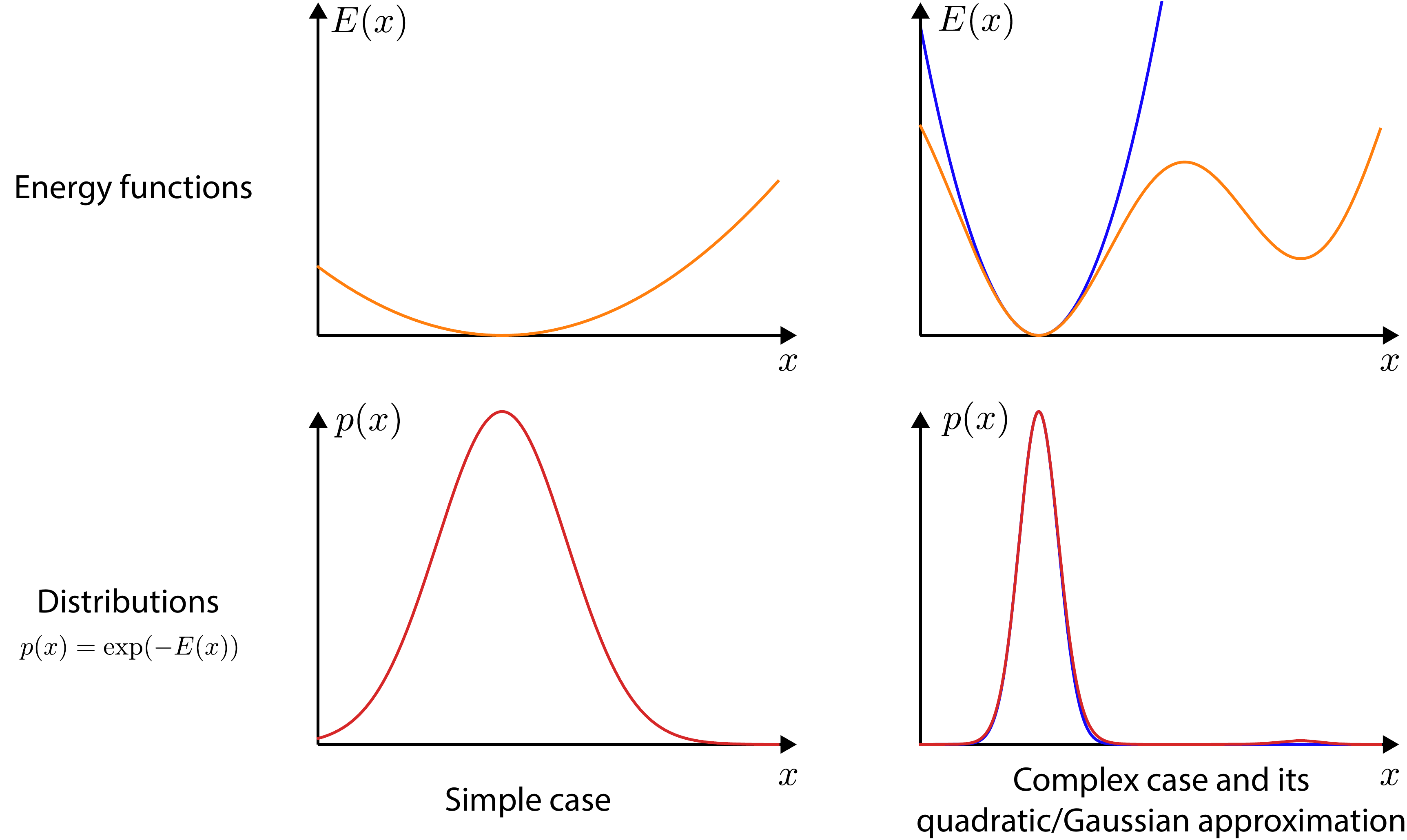}
	\caption{\textbf{Probabilistic interpretation.} Left: A quadratic energy function and its associated Gaussian error distribution. Right: A complex energy function, which is locally approximated with a Gaussian (blue) near the minimum. The curvature of the energy function is a measure of the confidence in the estimate and the variance of the associated error distribution. The energy on the right is more constrained and its error distribution has a lower variance.}
	\label{fig:energy}
\end{figure}

Our goal is to find the next best view for the drone at the future time step $t+1$, $\dronepose^{t+1}$.
We will model the uncertainty of the pose estimate in a probabilistic setting. Let $p(\pose|\openpose,\dronepose,\liftpose,\bone)$ be the posterior distribution of poses. Then, $E_\text{pose}$ is its negative logarithm and its minimization corresponds to maximum a posteriori (MAP) estimation. In this formalism, the sum of the individual terms in $E_\text{pose}$ models that our posterior distribution is composed of independent likelihood and prior distributions. For a purely quadratic term, $E(x) = \omega (x-\mu)^2$, the corresponding distribution $p_E = \exp{(-E)}$ is a Gaussian with mean $\mu$ and standard deviation $\sigma=\frac{1}{\sqrt{2 \omega}}$. Notably, $\sigma$ is directly linked to the weight $\omega$ of the energy. 
Most of our energy terms involve non-linear operations, such as perspective projection in $E_\text{proj}$, and therefore induce non-Gaussian distributions, as visualized in Fig.~\ref{fig:energy}.
Nevertheless, as for the simple quadratic case, the weights $\omega_p$ and $\omega_l$ of $E_\text{proj}$ and $E_\text{lift}$ can be interpreted as surrogates for the amount of measurement noise in the 2D and 3D pose estimates. 

\begin{figure}
	\centering
	\includegraphics[width=1\linewidth]{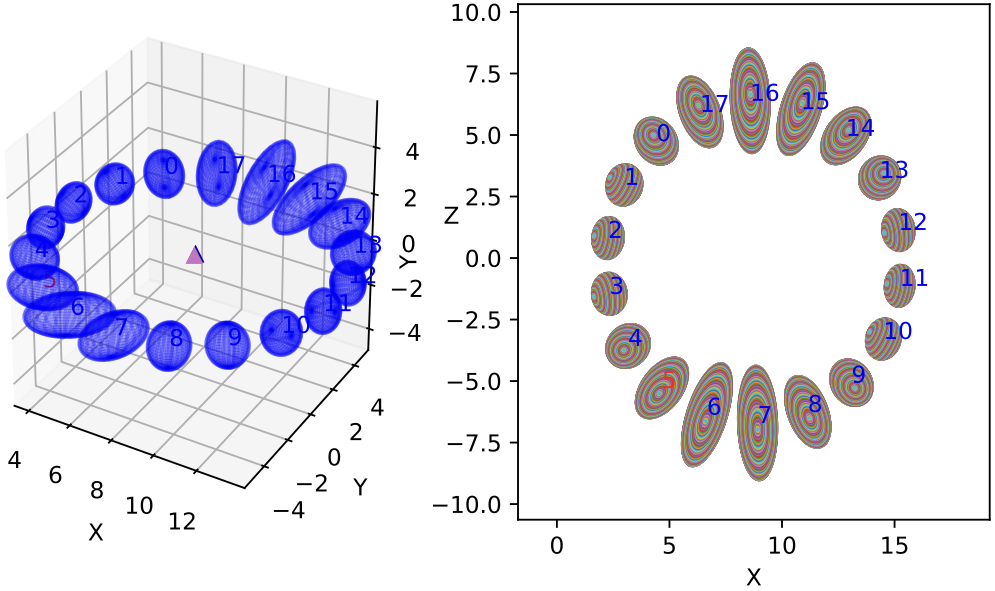}
	\caption{\textbf{Uncertainty estimates} for each candidate drone position, visualized on the left as 3D ellipsoids and on the right from a 2D top-down view. Each ellipse visualizes the eigenvalues of the hip location when incorporating an additional view from its displayed position. Here, the previous image was taken from the top (position 16) and uncertainty is minimized by moving to an orthogonal view. The complete distribution has more than three eigenvectors and cannot straightforwardly be visualized in 3D.}
	\label{fig: ellipse_fig}
\end{figure}

A good measure of uncertainty is the sum of the eigenvalues of the covariance $\Sigma_{p}$ of the underlying distribution $p$.
The sum of the eigenvalues captures the spread of a multivariate distribution with a single variable, similarly to the variance in the univariate case. To exploit this uncertainty estimation for our problem, we now extend $E_\text{pose}$ to model not only the current and past poses but also the future ones and condition it on the choice of the future drone position. 
To determine the best next drone pose, we sample candidate positions and chose the one with the lowest uncertainty. This process is illustrated in Figure~\ref{fig: ellipse_fig}.

\para{Future pose forecasting.}
In our setting, accounting for the dynamic motion of the person is key to successfully positioning the camera. We model the motion of the person from the current frame $t$ to the next $M$ future frames $t+i$, $i \in (1, \dots, M)$ linearly, i.e. we aim to keep the velocity of the joints constant across our window of frames. We  also constrain the future poses by the bone length term. The future pose vectors $\pose^{t+i}$ are constrained by the smoothness and bone length terms, but for now not by any image-based term since the future images are not yet available at time $t$. Minimizing this extended $E_\text{pose}$ for future poses gives the MAP poses $\hat{\pose}^{t+i}$. It continues the motion $\hat{\pose}^{t-k, \cdots, t+K}$ smoothly while maintaining the bone lengths.
As we predict only the near future, we have found this simple extrapolation to be sufficient. We leave as future work the use of more advanced methods~\cite{Fragkiadaki15,Zhang19d} to forecast further.

\para{Future measurement forecasting.}
We aim to find the future drone position, $\dronepose^{t+1}$, that reduces the posterior uncertainty, but we do not have footage from future viewpoints to condition the posterior on. Instead, we use the predicted future human pose $\hat{\pose}^{t+i}$, $i \in (1, \dots, M)$, as a proxy for $\liftpose^{t+i}$ and approximate $\openpose^{t+i}$ with the projection

\begin{equation}
\hat{\openpose}^{t+1} = \Pi(\hat{\pose}^{t+1}, \dronepose^{t+1}, \boldsymbol{K}) \;.
\end{equation}
At first glance, constraining the future pose on these virtual estimates in $E_\text{pose}$ does not add anything since the terms $E_\text{proj}$ and $E_\text{lift}$ are zero at $\hat{\pose}^{t+1}$ by this construction. However, it changes the energy landscape and models how strong a future observation would constrain the pose posterior. In particular, the projection term, $E_\text{proj}$, narrows down the solution space in the direction of the image plane but cannot constrain it in the depth direction, creating an elliptical uncertainty as visualized in Fig~\ref{fig: ellipse_fig}. The combined influence of all terms is conveniently modeled as the energy landscape of $E_\text{pose}$ and its corresponding posterior.

In our current implementation we assume that the 2D and 3D detections are affected by pose-independent noise, and their variance is captured by $\omega_p$ and $\omega_l$, respectively.
These factors could, in principle, be view dependent and in relation to the person's pose. For instance, \cite{Chao17} may be more accurate at reconstructing a front view than a side view.
However, while estimating the uncertainty in deep networks is an active research field \cite{Prokudin18}, predicting the expected uncertainty for an unobserved view has not yet been attempted for pose estimation. It is an interesting future work direction.

\para{Variance estimator.}
$E_\text{pose}$ and its corresponding posterior has a complex form due to the projection and prior terms. Hence, the sought-after covariance $\Sigma_p$ cannot be expressed in closed form and approximating it by sampling the space of all possible poses would be expensive. Instead, for the sake of uncertainty estimation, we approximate $p(\pose|\dronepose, \openpose, \liftpose,\bone)$ locally with a Gaussian distribution $q$, such that
\begin{align}
\Sigma_{p(\pose|\dronepose, \openpose, \liftpose)} \approx \Sigma_q \text{ where } q = N(\pose | \hat{\pose}, \Sigma_q) \;,
\end{align}
with $\hat{\pose}$ and $\Sigma_q$ the Gaussians mean and covariance matrix, respectively. Such an approximation is exemplified in Figure~\ref{fig:energy}.
For a Gaussian, the covariance of $q$ can be computed in closed form as the inverse of the Hessian of the negative log likelihood, $\Sigma_q = H^{-1}_{-\log{q}}$, where $H_{-\log{q}} = \frac{\partial^2 -\log{q(\pose)}}{\partial \pose}\Bigr|_{\pose = \hat{\pose}}$. 
Under the Gaussian assumption, $\Sigma_{p}$ is thereby well approximated by the second order gradients, $H^{-1}_{E_\text{pose}}$, of $E_\text{pose}$. 
Our experiments show that this simplification holds well for the introduced error terms.

To select the view with minimum uncertainty among a set of $K$ candidate drone trajectories, we therefore
\begin{enumerate}
\item optimize $E_\text{pose}$ once to forecast $M$ human poses $\hat{\pose}^{t+i}$, for $1 \leq i \leq M$\item use these forecasted poses to set $\hat{\liftpose}^{t+i}$ and $\hat{\openpose}^{t+i}$ for each $1 \leq i \leq M$ for each candidate trajectory $c$,
\item compute the second order derivatives of $E_\text{pose}$ for each $c$, which form $H_c$, and
\item compute and sum up the respective eigenvalues to select the candidate with the least uncertainty.
\end{enumerate}

	\para{Discussion.}
	In principle, $p(\pose|\openpose,\dronepose,\liftpose,\bone)$, i.e. the probability of the most likely pose, could also act as a measure of certainty, as implicitly used in~\cite{Rahimian16} on a known motion trajectory to minimize triangulation error. However, the term $E_\text{proj}(\hat{\pose},\hat{\openpose})$ of $E_\text{pose}$ is zero for the future time step $t+i$, because the projection of $\hat{\pose}^{t+i}$ is by construction equal to $\hat{\openpose}^{t+i}$ and therefore uninformative.
	Another alternative that has been proposed in the literature is to approximate the covariance through first order estimates \cite{Tkach17}, as a function of the Jakobi matrix. However, as also the first order gradients of $E_\text{proj}$ vanish at the MAP estimate, this approximation is not possible in our case.

\subsection{Drone Control Policies and Flight Model}\label{sec:control_policies}

In the experiments where we simulate drone flight, the algorithm decides between $9$ candidate trajectories in the directions up, down, left, right, up-right, up-left, down-right, down-left and center. To ensure that the drone stays a fixed distance away from the person, the direction vector is normalized by the fixed-distance value. 
	
In the remainder of this section, we describe how we model the flight of the drone so that we can predict the position of the drone along a potential trajectory in future time steps. By forecasting the future $M$ locations of the drone on a potential trajectory $c$, we can predict the 2D pose estimations $\hat{\openpose}^{t+i}$ for each $\{i\}_{i=1}^M$  more accurately.

We control the flight of our drone by passing it the desired velocity vector and the desired yaw rotation amount with the maximum speed kept constant at $5$ m/s. The drone is sent new commands once every $\Delta t=0.2$ seconds. 

We model the drone flight in the following manner. We assume that the drone moves with constant acceleration during a time step $\Delta t$. If the drone has current position $x_\text{current}$ and velocity $V_\text{current}$, then with an current acceleration $a_\text{current}$, its next position $x_\text{goal}$ in $\Delta t$ time will be
\begin{equation}
x_\text{goal} = x_\text{current} + V_\text{current}\Delta t + 0.5 a_\text{current} \Delta t^2\;.
\label{eq:motion_equation}
\end{equation}

The current acceleration at time $t$ is found as a weighted average of the input acceleration $a_\text{input}$ and the acceleration of the previous step  $a_\text{previous}$. This can be written as
\begin{equation}
a_\text{current} = \alpha a_\text{input} + (1-\alpha) a_\text{previous}.
\label{eq:acceleration}
\end{equation}

 $a_\text{input}$ is determined according to the candidate trajectory being evaluated. The direction of the acceleration vector is set to the direction of the candidate trajectory. We determine the magnitude of the input acceleration through least-square minimization of the difference between the predicted $x_\text{goal}$ and the actual drone position. $\alpha$ is found by line search.
  
By estimating the future positions of the drone, we are able to forecast more accurate future 2D pose estimations, leading to more accurate decision making. Examples of predicted trajectories are shown in Figure~\ref{fig:predicted_traj}. Further details are provided in the supplementary material.

\begin{figure}
	\centering
	\includegraphics[width=0.47\textwidth]{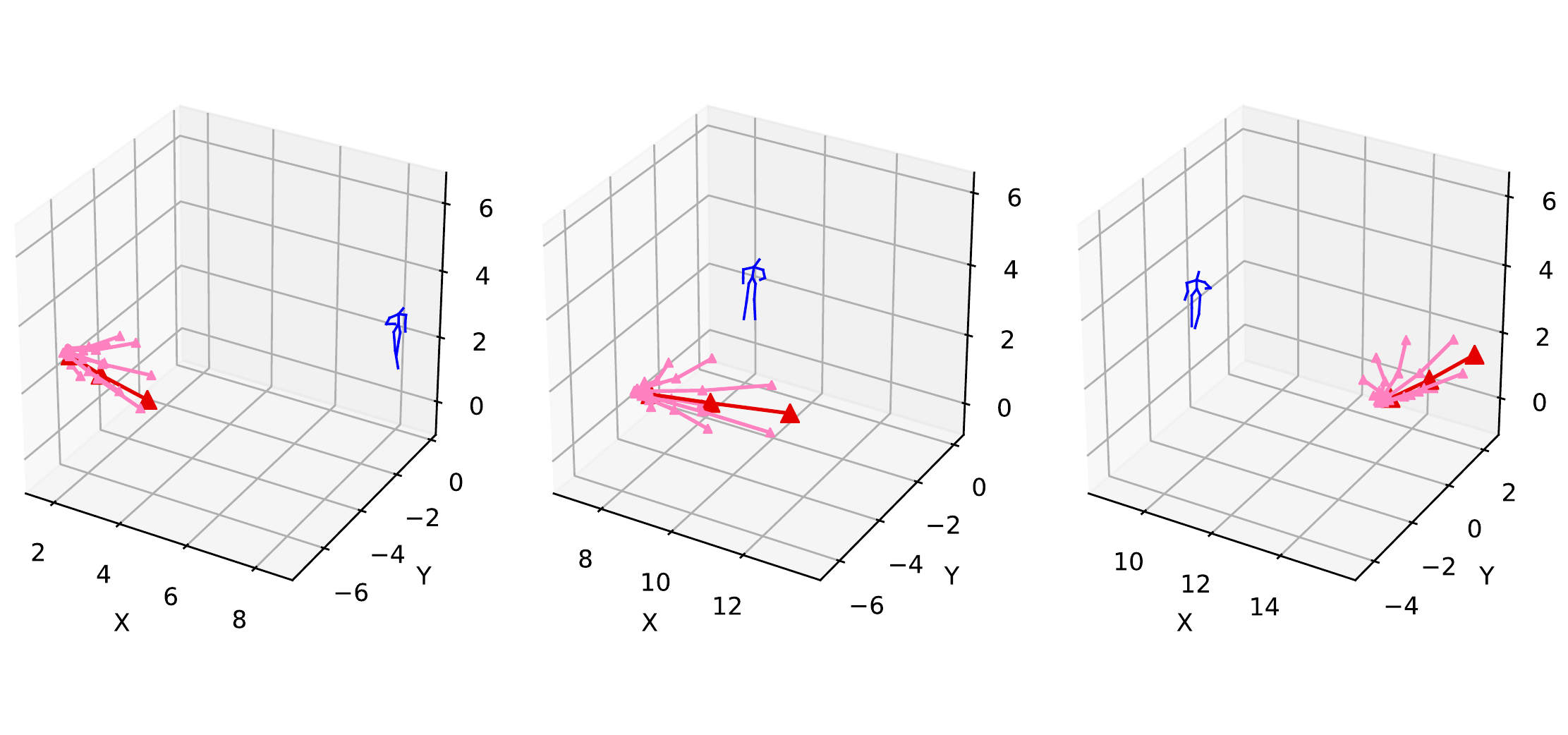}
	\caption{\textbf{Predicted trajectories} as the drone is circling the subject. The future drone positions are predicted for the future $3$ steps, represented by triangle markers on the trajectories. Red depicts the chosen trajectory.}
	\label{fig:predicted_traj}
\end{figure}

\section{Evaluation}

In this section we evaluate the improvement on 3D human pose estimation that is achieved through optimization of the drone flight.

\para{Simulation environment.} Although \cite{Redmon18, Cao17,Tekin17a} run in real time, and online SLAM from a monocular camera \cite{Davison07} is possible, we use a drone simulator since the integration of all components onto constrained drone hardware is difficult and beyond our expertise. 
We make simulation realistic by driving our characters with real motion capture data from the CMU Graphics Lab Motion Capture Database \cite{CMUHMC} and using the AirSim~\cite{Shah17} drone simulator that builds upon the Unreal game engine and therefore produces realistic images of natural environments. Simulation also has the advantage that the same experiment can be repeated with different parameters and be directly compared to baseline methods and ground-truth motion.

\para{Simulated test set.}
We test our approach on three CMU motions of increasing difficulty: 
\emph{Walking} straight (subject 2, trial 1), 
\emph{Dance} with twirling (subject 5, trial 8), and
\emph{Running} in a circle (subject 38, trial 3). %
Additionally, we use a validation set consisting of \emph{Basketball} dribble (subject 6, trial 13), and
 \emph{Sitting} on a stool (subject 13, trial 6), to conduct a grid search for hyperparameters.

\begin{figure}
	\centering
	\includegraphics[width=0.4\textwidth]{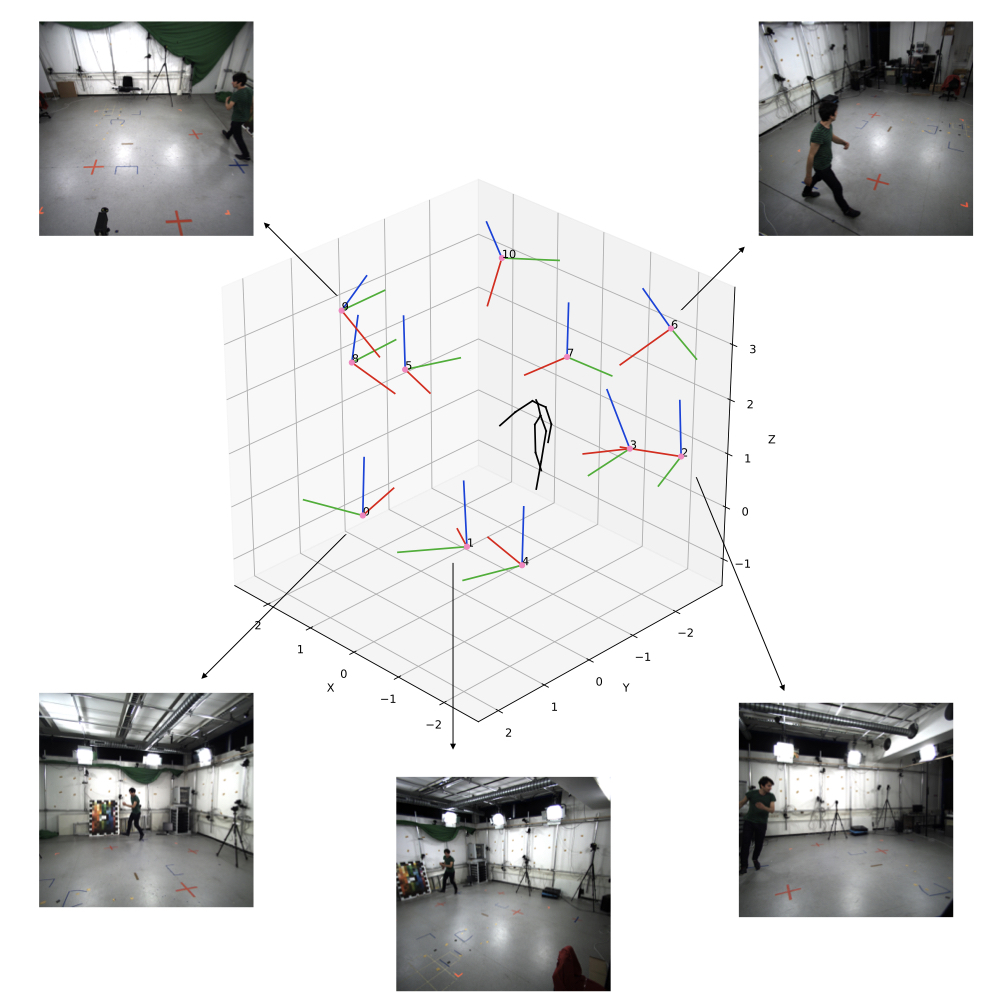}
	
	\caption{\textbf{MPI\_INF\_3DHP dataset}, which has images taken from $14$ viewpoints with various distances to the subject. We use this dataset to evaluate our performance on datasets with realistic camera positioning and real images.}
	\label{fig:mpi_dataset}
\end{figure}

\para{Real test set.}
To show that our planner also works outside the simulator, we evaluate our approach on a section of the MPI-INF-3DHP dataset, which includes motions such as running around in a circle and waving arms in the air. The dataset provides $14$ fixed viewpoints that are at varying distances from one another and from the subject, as depicted in Figure~\ref{fig:mpi_dataset}. In this case, the best next view is restricted to one of the $14$ fixed viewpoints. This dataset lets us evaluate whether the object detector of~\cite{Redmon18}, the 2D pose estimation method of~\cite{Chao17}, and the 3D pose regression technique of~\cite{Tekin17a} are reliable enough in real environments. Since we cannot control the camera in this setting, we remove those cameras from the candidate locations where we predict that the subject will be out of the viewpoint.

\begin{figure}
	\centering
	\includegraphics[width=0.45\textwidth]{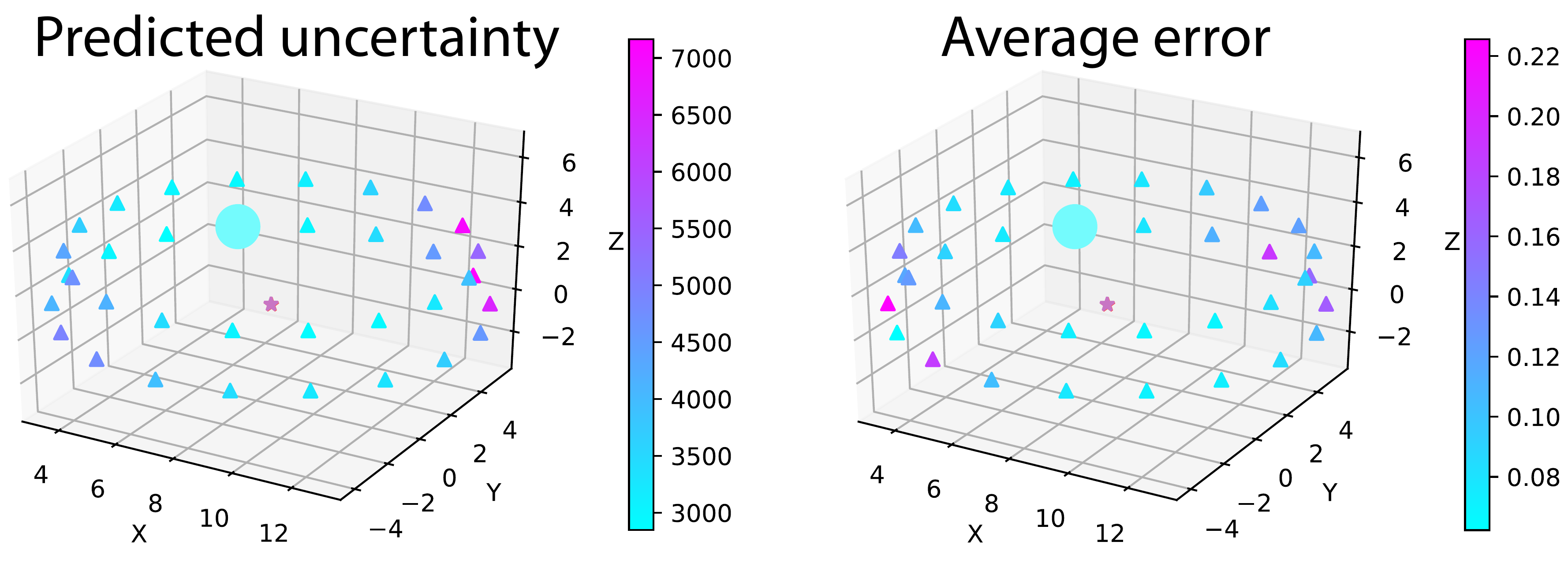}
	\caption{\textbf{Uncertainties estimates} across potential viewpoints (left image)  compared with the average error we would obtain if we were to visit these locations (right image). The star represents the location of the subject and the large circle depicts the chosen viewpoint according to the lowest uncertainty.}
	\label{fig:uncertainty_error}
\end{figure}

\begin{table*}[htbp]
\centering
\resizebox{0.7\linewidth}{!}{
\begin{tabular}{|l|c|c|c|c|c||c|}

\hline   					&	\multicolumn{4}{c|}{Noisy ground truth} & Networks & \\ 
\hline                      & CMU-Walk & CMU-Dance & CMU-Run & MPI-INF-3DHP & MPI-INF-3DHP  & Total \\ 
\hline
Oracle     					& 0.101$\pm$0.001 & 0.101$\pm$0.001 & 0.109$\pm$0.001 &  0.136$\pm$0.002 & 0.17$\pm$0.0005 & 0.142$\pm$0.027 \\

Ours (Active)          & \textbf{0.113}$\pm$0.001 & \textbf{0.116}$\pm$0.003 & \textbf{0.135}$\pm$0.002 & \textbf{0.145}$\pm$0.006 & \textbf{0.21}$\pm$0.0008 &  \textbf{0.144}$\pm$0.35 \\
Random                  & 0.123$\pm$0.002&0.125$\pm$0.003 &0.159$\pm$0.003& 0.286$\pm$0.027& 0.28$\pm$0.03  &  0.195$\pm$0.07\\
Constant Rotation   & 0.157$\pm$0.002&0.146$\pm$0.004&0.223$\pm$0.003&0.265$\pm$0.010& 0.29$\pm$0.03  & 0.216$\pm$0.06 \\
Constant Angle       & 0.895$\pm$0.54 &0.683$\pm$0.31&0.985$\pm$0.24&1.45$\pm$0.63& 1.73$\pm$0.61  & 1.15$\pm$0.38 \\ \hline
\end{tabular}}
\smallskip
\caption{\textbf{3D pose accuracy on the teleportation experiment}, using noisy ground truth to estimate $\openpose$ and $\liftpose$ in the first three columns, and using the networks of \cite{Zhao17, Tekin17a} in the fourth column. We outperform all predefined baseline trajectories and approach the accuracy of the oracle that has access to the average error of each candidate position.}
\label{tab:toy_online_mode_gt}
\end{table*}

\para{Baselines.}
Existing drone-based pose estimation methods use predefined policies to control the drone position relative to the human. Either the human is followed from a constant angle and the angle is set externally by the user~\cite{Naegeli18} or the drone undergoes a constant rotation around the human~\cite{Zhou18}. As another baseline, we use a random decision policy, where the drone picks uniformly randomly among the proposed viewpoints. Finally, the oracle is obtained by moving the drone to the viewpoint where the reconstruction in the next time step will have the lowest average error, which is achieved by exhaustively trying all 
viewpoints \emph{with} the corresponding image in the next time frame. 

\para{Hyper parameters.} We set the weights of the loss term for the reconstruction as follows: $\omega_p=0.0001$ (projection), $\omega_s=1$ (smoothness), $\omega_l=0.1$ (lift term), $\omega_b=1$ (bone length), which were found by grid search. We set the weights for the decision making as $\omega_p=0.001$, $\omega_s=1$, $\omega_l=0.1$, $\omega_b=1$ . Our reasoning is, we need to set the weights of the projection and lift terms slightly lower because they are estimated with large noise, which is introduced by the neural networks or as additive noise. However, they do not need to be as low for the uncertainty estimation. 

\subsection{Analyzing Reconstruction Accuracy}
We report the mean Euclidean distance per joint in meters in the middle frame of the temporal window we optimize over. For teleportation mode, the size of the temporal window is set to $k=2$ past frames and $1$ future frame, and for the drone flight simulations, to $k=6$ for past frames and $3$ future frames. 

\para{Simulation Initialization.} 
The frames are initialized by \emph{back-projecting} the 2D joint locations estimated in the first frame, $\openpose^{t=0}$, to a distance $d$ from the camera that is chosen such that the back-projected bone lengths match with the average human height. We then refine this initialization by running the optimization without the smoothness term, as there is only one frame. All the sequences are evaluated for $120$ frames, with the animation sequences played at $5$ Hz.

\para{Teleportation Mode.}
To understand whether our uncertainty predictions for potential viewpoints coincide with the actual 3D pose errors we will have at these locations, we run the following simulation: We sample a total of $18$ points on a ring around the person, as shown in Fig.~\ref{fig:uncertainty_error}, and allow the drone to teleport to any of these points. We optimize over a total of $k=2$ past frames and forecast $1$ frame into the future. We chose this window size to emphasize the importance of the next choice of frame.

\begin{figure}[t!]
    \centering
    \includegraphics[width=0.35\textwidth]{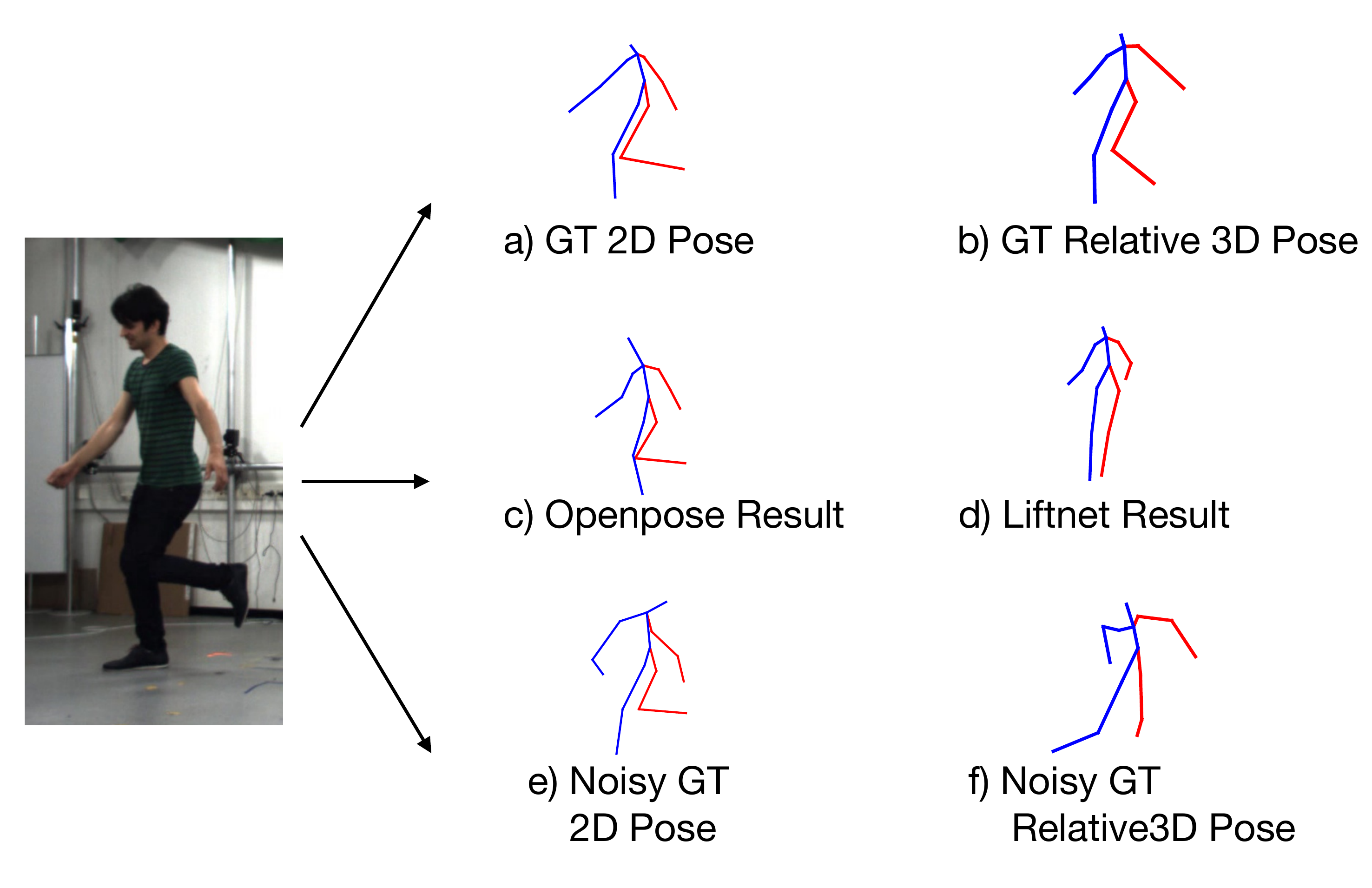}
    \caption{\textbf{Example image from the MPI-INF-3DHP dataset} along with the 2D pose detections $\openpose$ and 3D relative pose detections $\liftpose$ obtained using ground truth, noisy ground truth or the networks of \cite{Cao17} and \cite{Tekin17a}. The noise we add on the ground truth poses is determined according to the statistics of \cite{Cao17} and \cite{Tekin17a}, measured on our validation set.}
    \label{fig:openpose_liftnet_noise}
\end{figure}

We perform two variants of this experiment. In the first one, we simulate the 2D and 3D pose estimates, $\openpose,\liftpose$, by adding Gaussian noise to the ground-truth data. The mean and standard deviation of this noise is set as the error of  \cite{Cao17} and \cite{Tekin17a}, run on the validation set of animations. Figure~\ref{fig:openpose_liftnet_noise} shows a comparison between the ground truth values, noisy ground truth values and the network results.  The results of this experiment are reported in Table~\ref{tab:toy_online_mode_gt}, where we also provide the standard deviations across 5 trials with varying noise and starting from different viewpoints. On the MPI-INF-3DHP dataset, we also provide results using~\cite{Cao17} and~\cite{Tekin17a} on the simulator images to obtain the 2D and 3D pose estimates. Further results are in the supplementary material.

\begin{figure}
	\centering
	\includegraphics[width=0.48\textwidth]{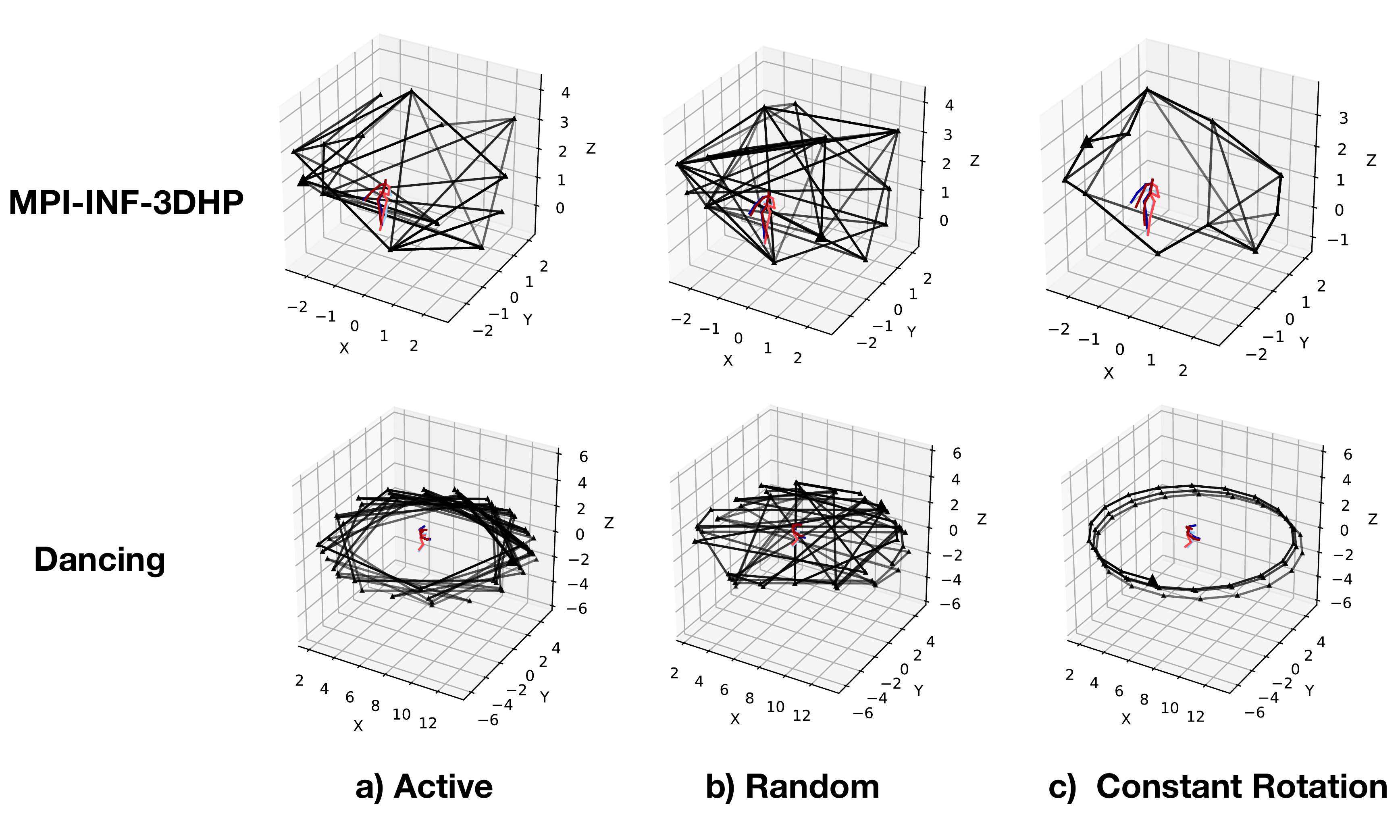}
	\caption{\textbf{Trajectories found by our active planner} along with random and constant rotation baselines. The first row depicts the trajectories for the MPI-INF-3DHP dataset, and the second row shows the trajectories for the dancing motion. The trajectories obtained with our algorithm are regular and look different from the random trajectories, especially for the dancing motion. Our algorithm prefers trajectories resulting in large angular variance with respect to the subject between viewpoints.}
	\label{fig:trajectories_toy}
\end{figure}

Altogether, the results show that our active motion planner achieves consistently lower error values than the baselines and we come the closest to achieving the best possible error for these sequences and viewpoints, despite having no access to the true error. The random baseline also performs quite well in these experiments, as it takes advantage of the drone teleporting to a varied set of viewpoints. The trajectories generated by our active planner and the baselines is depicted in Figure ~\ref{fig:trajectories_toy}. Importantly, Figure~\ref{fig:uncertainty_error} evidences that our predicted uncertainties accurately reflect the true pose errors, thus making them well suited to our goal.

\para{Simulating Drone Flight.}
To evaluate more realistic cases where the drone is actively controlled and constrained to only move to nearby locations, we simulate the drone flight using the AirSim environment.
While simulating drone flight, we target a fixed radius of $7$m from the subject and therefore provide direction candidates that lead to preserving this distance. We do not provide samples at different distances, as moving closer is unsafe and moving farther leads to more concentrated image projections and thus higher 3D errors.
We also restrict the drone from flying outside the altitude range $0.25$m-$3.5$m, so as to avoid crashing into the ground and flying above the subject.

In this set of experiments, we \emph{fly} the drone using the simulator's realistic physics engine. To this end, we sample $9$ candidate directions towards up, down, left, right, up-right, up-left, down-right, down-left and center. We then predict the $3$ consecutive future locations using our simplified (closed form) physics model, to get and estimate where the drone will be at when continuing in each of the $9$ directions. We then estimate the uncertainty at these sampled viewpoints and choose the minimum.

\begin{table}[t]
\centering
\resizebox{1.01\linewidth}{!}{
\begin{tabular}{|l|l|l|l||l|}
\hline  						&CMU-Walk & CMU-Dance & CMU-Run & Total \\ \hline
Ours (Active)    		 & \textbf{0.26}$\pm$0.03 & 0.22$\pm$0.04 & 0.44$\pm$0.04 & 0.31$\pm$0.10\\
Constant Rotation  	& 0.28$\pm$0.06 & \textbf{0.21}$\pm$0.04 & \textbf{0.41}$\pm$0.02 & \textbf{0.30}$\pm$0.08\\ 
Random 					& 0.60$\pm$0.13 & 0.44$\pm$0.19 & 0.81$\pm$0.16 & 0.62$\pm$0.15\\ 
Constant Angle   	& 0.41$\pm$0.07 & 0.63$\pm$ 0.06 & 1.26$\pm$0.17 & 0.77$\pm$0.36\\ \hline
\end{tabular}}
\smallskip
\caption{\textbf{Results of drone full flight simulation}, using noisy ground truth as input to estimate $\openpose$ and $\liftpose$. The results of constant rotation are the average of $10$ runs, with $5$ runs rotating clockwise and $5$ counter-clockwise. Our approach yields results comparable to those of constant rotation, outperforming the other baselines. The trajectory our algorithm draws also results in a constant rotation, the only difference being the rotation direction.}
\label{tab:flight_openpose}
\end{table}

\begin{figure}
	\centering
	\includegraphics[width=0.47\textwidth]{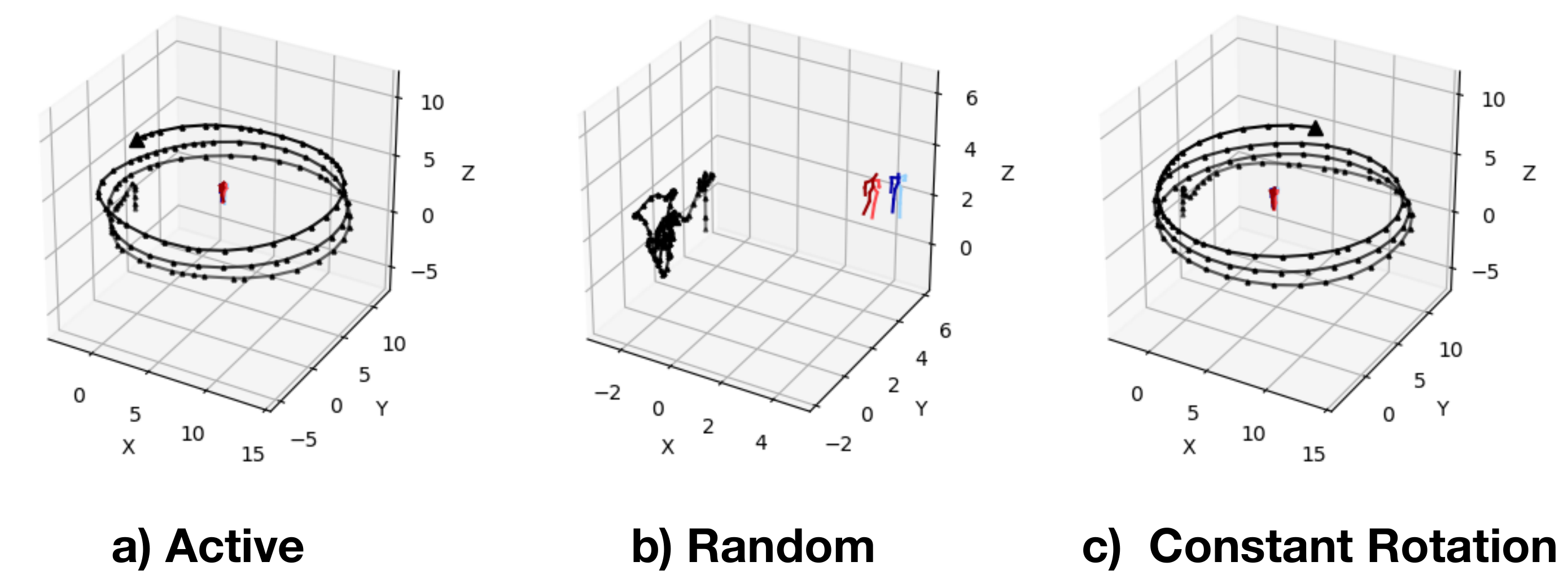}
	\caption{\textbf{Trajectories found during flight} by our active planner and the baselines. Our algorithm also chose to perform constant rotation. Because of the drone momentum, the random baseline cannot increase the distance between its camera viewpoints.}
	\label{fig:trajectories_flight}
\end{figure}

We achieve comparable results to constant rotation on simulated drone flight. In fact, except for the first few frames where the drone starts flying, we observe the same trajectory as constant rotation, only the rotation direction varies. Constant rotation being optimal in this setting is not counter-intuitive, as constant rotation is very useful for preserving momentum. This allows the drone to sample viewpoints as far apart from one another as possible, while keeping the subject in view. Figure ~\ref{fig:trajectories_flight} depicts the different baseline trajectories and the active trajectory.

\section{Conclusion and Future Work}

 We have proposed a theoretical framework for estimating the uncertainty of future measurements from a drone. This permits us to improve 3D human pose estimation by optimizing the drone flight to visit those locations with the lowest expected uncertainty. We have demonstrated with increasingly complex examples, in simulation with synthetic and real footage, that this theory translates to closed-loop drone control and improves pose estimation accuracy. We envision our approach being developed further for improving the performance of athletes and performance artists. It is important to preserve the subjects' privacy in such autonomous systems. We encourage researchers to be sensitive to this issue.
 
Key to the success of our approach is the integration of several sources of uncertainty. Our primary goal was to make uncertainty estimation tractable, but further improvements are needed to run it on an embedded drone system. The current implementation runs at $0.1$Hz, but the optimization is implemented in Python using the convenient but slow automatic differentiation of PyTorch to obtain second derivatives. Furthermore, we have considered a physically plausible drone model but neglected physical obstacles and virtual no-go areas that would restrict the possible flight trajectories. In the case of complex scenes with dynamic obstacles, we expect our algorithm to outperform any simple, predefined policy. Currently, we assume a constant error for the 2D and 3D pose estimates. In future work, we will investigate how to derive situation-dependent noise models of deep neural networks. Furthermore, we plan to study new ways of estimating the uncertainty of the deployed deep learning methods and extend our work to optimize drone trajectories for different computer vision tasks. 
 
\section{Acknowledgements}

This work was supported in part by the Swiss National Science Foundation and by a Microsoft Joint Research
Project.
% !TEX root = ../top.tex
% !TEX spellcheck = en-US

\section{Supplementary Material}

\subsection{Supplementary Video}

The supplementary video provides a short overview of our work and summarizes the methodology and results. It includes video results of our active trajectories for both the teleportation and simulated flight cases.

\subsection{Drone Flight Simulation}

We mention in Section~$4.1$ of our main document that we obtain the constant rotation trajectory on simulated drone flight, albeit with varying rotation direction. We report the results of constant rotation in both directions in Table~\ref{tab:rotation_both_dir}, along with our active trajectory’s results. The results show that in general, the active trajectory’s error values are in be- tween the constant rotation to the right and the left. This is because the active trajectory’s direction varies, however the trajectory is equivalent to constant rotation.

% !TEX root = ../top.tex
% !TEX spellcheck = en-US

\begin{table}[htbp]
	\centering
	\resizebox{1\linewidth}{!}{
		\begin{tabular}{|l|c|c|c|c||c|}
			\hline                        &CMU-Walk & CMU-Dance & CMU-Run & Total\\ \hline
			
			Ours (Active)     & 0.26$\pm$0.03 & 0.22$\pm$0.04 &0.44$\pm$0.04 & 0.31$\pm$0.10  \\ 
			Constant rot. (CW)       & 0.22$\pm$0.004 & 0.18$\pm$0.01 & 0.41$\pm$0.02 & 0.27$\pm$0.10  \\
			Constant rot. (CCW)     & 0.35$\pm$0.01 &0.24$\pm$0.04 & 0.41$\pm$0.02 & 0.34$\pm$0.08 \\ \hline
	\end{tabular}}
	\caption{{\bf Results of drone full flight simulation}, as Table~2 of our main document, reporting the error of constant rotation in both directions. ”CW” and ”CCW” stand for ”clock-wise” and ”counter clock-wise” respectively. In general, the active trajectory’s error values are in between the error values of constant rotation in the right and left directions.}
	\label{tab:rotation_both_dir}
\end{table}

\subsection{The Drone Flight Model}

As we mention in Section~$3.3$ of our main document, in order to accurately predict where the drone will be positioned after passing it a goal velocity, we have formulated a drone flight model.

\begin{figure}[h]
	\centering
	\includegraphics[width=0.4\textwidth]{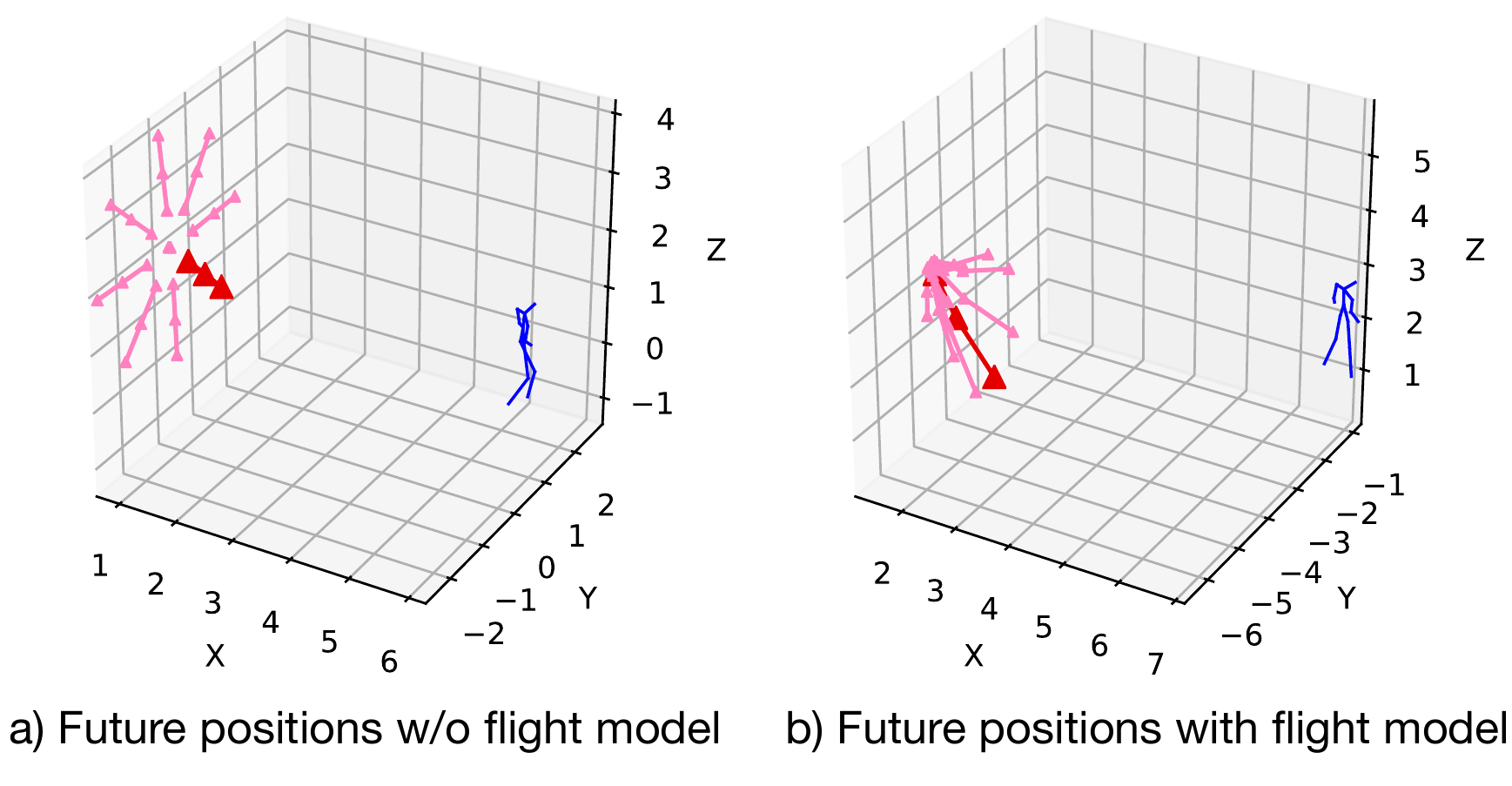}
	\caption{\textbf{The predicted future positions of the drone} (a) without using our flight model and (b) using our flight model.}
	\label{fig:trajectories_ablation}
\end{figure}

\begin{figure}
	\centering
	\includegraphics[width=0.4\textwidth]{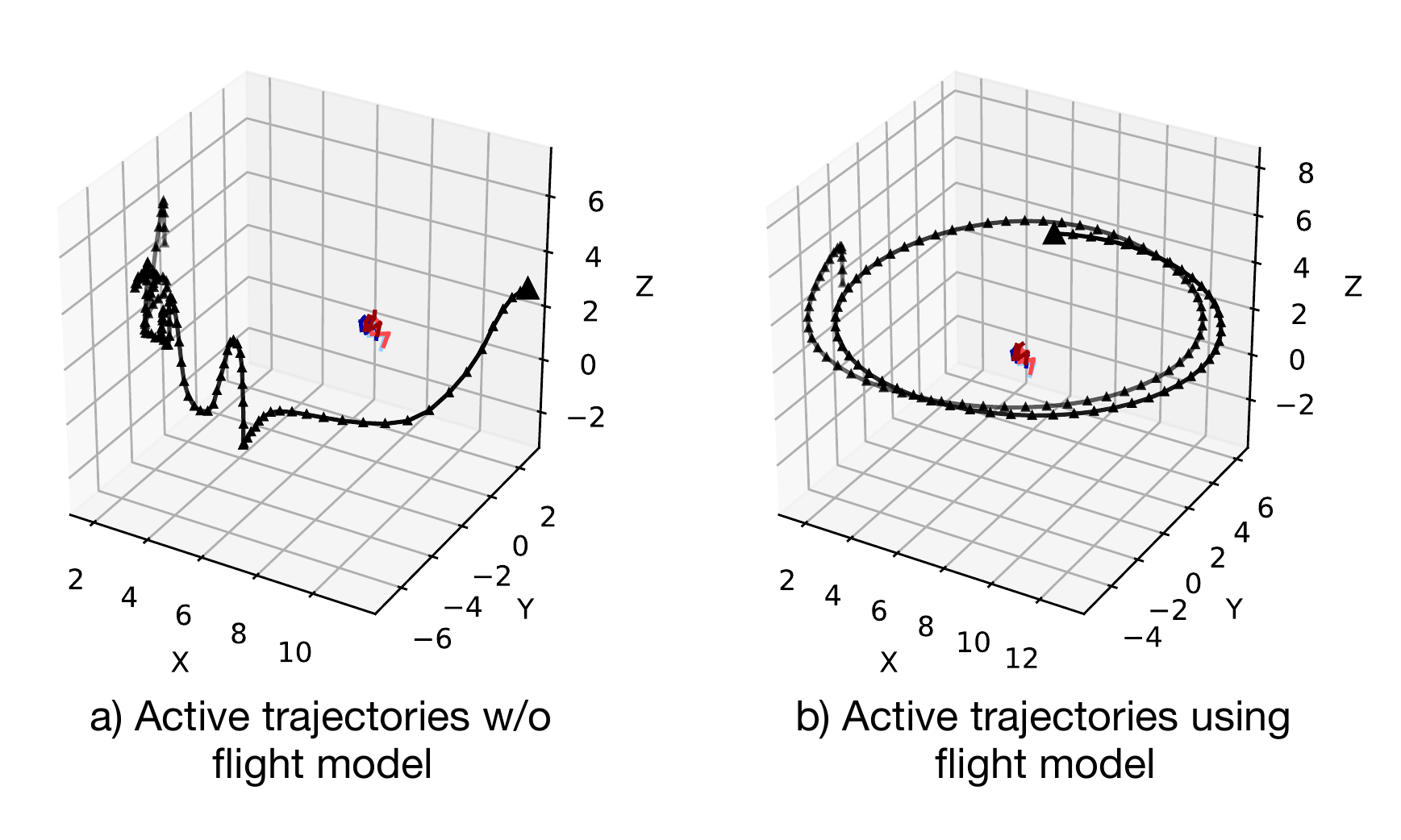}
	\caption{\textbf{The trajectories drawn by our active decision making policy} (a) without using our flight model and (b) using our flight model. We are able to find the well performing policy of constant rotation when we are using more realistic sampling of future drone positions, found using our drone flight model.}
	\label{fig:trajectories_ablation2}
\end{figure}

\para{Ablation Study.} We replace our drone flight model with uniform sampling around the drone. This is illustrated in Figure~\ref{fig:trajectories_ablation}. We evaluate the performance of our active decision making policy with the uniform sampling in Table~\ref{tab:ablation}. The trajectories found using this sampling policy is shown in Figure~\ref{fig:trajectories_ablation2}. We find that the algorithm cannot find the constant rotation policy when we remove the drone flight model and in turn, performs worse.

% !TEX root = ../top.tex
% !TEX spellcheck = en-US

\begin{table}[htbp]
	\centering
	\resizebox{1\linewidth}{!}{
		\begin{tabular}{|l|c|c|c|c||c|}
			\hline                        &CMU-Dribble & CMU-Sitting & CMU-Dinosour & Total\\ \hline
			
			Active with Flight Model      & \textbf{0.28}$\pm$0.006 & \textbf{0.15}$\pm$0.007 &\textbf{0.12}$\pm$0.02 & \textbf{0.18}$\pm$0.01  \\ 
			Active w/o Flight Model       & 0.65$\pm$0.09 & 0.48$\pm$0.09 & 0.22$\pm$0.07 & 0.45$\pm$0.08  \\
			Constant Rot.                 & 0.30$\pm$0.02 &\textbf{0.15}$\pm$0.01 & 0.15$\pm$0.03 & 0.20$\pm$0.02 \\ \hline
	\end{tabular}}
	\caption{{\bf Ablation study on the importance of having a drone flight model}. We show 3D pose accuracy on simulated drone flight using noisy ground truth for estimating $\openpose$ and $\liftpose$. We show that we have a large improvement when we use our flight model to predict the future locations of the drone. Using a flight model allows us to find the same trajectories as constant rotation.}
	\label{tab:ablation}
\end{table}

% !TEX root = ../top.tex
% !TEX spellcheck = en-US

% !TEX root = ../top.tex
% !TEX spellcheck = en-US

\subsection{Results with Openpose and Liftnet}

We evaluate our results on the toy example case, using the networks of \cite{Cao17} and \cite{Tekin17a} to find the 2D pose detections $\openpose$ and 3D relative pose detections $\liftpose$. The results are reported in Table~\ref{tab:toy_online_mode_with_openpose}. We outperform the baselines significantly for the real image dataset MPI-INF-3DHP. For the synthetic images, somes we are outperformed by random, but its error has much higher standard deviation and the difference between ours and random is within 1 standard deviation. 

We outperform the baselines significantly in the real image dataset as compared to the synthetic datasets because the error of network \cite{Cao17} for real data is much lower than for synthetic data. We verify this by comparing the normalized 2D-pose estimation errors of a synthetic sequence and a sequence taken from the MPI-INF-3DHP dataset. We find that the normalized average error of \cite{Cao17} of the synthetic sequence is $0.10$ with $0.08$ standard deviation, whereas the normalized average error of the real image sequence is $0.06$ with $0.06$ standard deviation. Therefore, the unrealistically high noise of OpenPose on the synthetic data deprives strong conclusions from the first three columns of Table~\ref{tab:toy_online_mode_with_openpose}. 

Oracle still performs very well for synthetic images in this case, but oracle makes decisions knowing the results of \cite{Cao17} for all candidate locations. However, this is impossible in practice due to the inherent uncertainty.

When the 2D pose detector is not unreliable, as in the case of Table~$1$ of our main document, we outperform random on all cases, well outside $2$ standard deviations.

For the case of the MPI-INF-3DHP dataset, we remove the $4$ ceiling cameras for this set of experiments. Since the networks of \cite{Cao17} and \cite{Tekin17a} were not trained with views from such angles they give highly noisy results which would also add noise to the values we report. 

% !TEX root = ../top.tex
% !TEX spellcheck = en-US

\begin{table}[htbp]
	\centering
	\resizebox{1\linewidth}{!}{
		\begin{tabular}{|l|c|c|c|c||c|}
			\hline              &CMU-Walk & CMU-Dance & CMU-Run & MPI-INF-3DHP. & Total\\ \hline
			Oracle              & 0.13$\pm$0 & 0.15$\pm$0 & 0.16$\pm$0.0005& 0.17$\pm$0.0005 & 0.15$\pm$0.0003 \\
			
			Ours (Active)      & \textbf{0.16}$\pm$0.005 & 0.25$\pm$0.0009 & 0.25$\pm$0.002 & \textbf{0.21}$\pm$0.0008 & \textbf{0.22}$\pm$0.002 \\ \hline
			Random             & 0.17$\pm$0.004 & \textbf{0.24}$\pm$0.01 &  \textbf{0.24}$\pm$0.005 & 0.28$\pm$0.03 & 0.23$\pm$0.01 \\
			Constant Rot.      & 0.20$\pm$0.002 & 0.28$\pm$0.02 & 0.28$\pm$0.001 & 0.29$\pm$0.007 & 0.26$\pm$0.007 \\
			Constant Angle     & 0.71$\pm$0.50 & 0.76$\pm$0.37 & 0.69$\pm$0.22 & 1.26$\pm$0.53 & 0.72$\pm$0.4 \\ \hline
	\end{tabular}}
	\caption{{\bf3D pose accuracy on toy experiment}, using \cite{Cao17, Tekin17a} for estimating $\openpose$ and $\liftpose$. We outperform all predefined baseline trajectories for the real image dataset, MPI-INF-3DHP. As for the cases with synthetic input, we achieve comparable results with random, albeit with much lower standard deviation.}
	\label{tab:toy_online_mode_with_openpose}
\end{table}

\subsection{Further Details About Simulation Environment}
\begin{figure}
	\centering
	\includegraphics[width=1\linewidth]{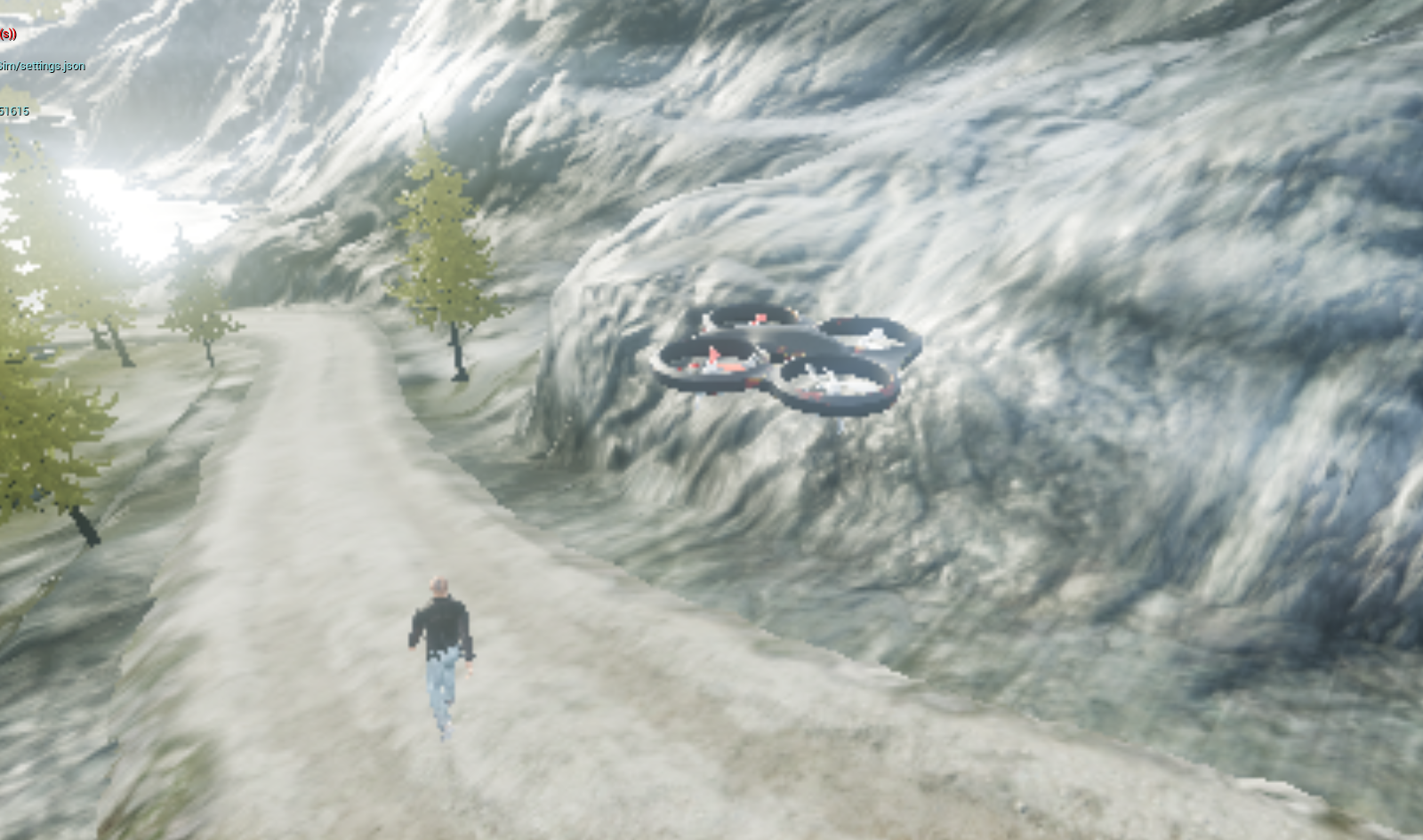}
	\caption{\textbf{Image of the simulation environment}, AirSim.}
	\label{fig:airsim}
\end{figure}

To test our algorithms we use the AirSim \cite{Shah17} drone simulator, a plug-in built for the Unreal game engine. An image from the simulator is shown in Figure~\ref{fig:airsim}.

% !TEX root = ../top.tex
% !TEX spellcheck = en-US

AirSim provides a Python API which can be used to control the drone realistically, since it uses the same flight controllers as used on actual drones. The position and orientation of the drone can be retrieved from the simulator according to the world coordinate system, which takes the drone's starting point as the origin. The drone can be commanded to move to a with a specified velocity for a specified duration. We have added functionalities to the simulator to control a human character, get ground truth information about the character and animate it with motions from the CMU Graphics Lab Motion Capture Database \cite{CMUHMC}.

For experiments requiring teleportation we use the simulator in "ComputerVision" mode, whereas for experiments simulating flight we use "Multirotor" mode.

{\small
\bibliographystyle{ieee}
\bibliography{string,vision,graphics,learning,robotics}
}

\end{document}